# Incremental compilation of Bayesian networks

M. Julia Flores
Department of Computer Science.
University of Castilla-La Mancha
Campus Universitario s/n.
Albacete. 02071 Spain

José A. Gámez
Department of Computer Science.
University of Castilla-La Mancha
Campus Universitario s/n.
Albacete. 02071 Spain

Kristian G. Olesen
Department of Computer Science.
Aalborg University
Frederik Bajers Vej 7E
DK-9220 Aalborg Øst. Denmark

## Abstract

Most methods for exact probability propagation in Bayesian networks do not carry out the inference directly over the network, but over a secondary structure known as a junction tree or a join tree (JT). The process of obtaining a JT is usually termed *compilation*. As compilation is usually viewed as a whole process; each time the network is modified, a new compilation process has to be performed. The possibility of reusing an already existing JT in order to obtain the new one regarding only the modifications in the network has received only little attention in the literature. In this paper we present a method for incremental compilation of a Bayesian network, following the classical scheme in which triangulation plays the key role. In order to perform incremental compilation we propose to recompile only those parts of the JT which may have been affected by the network's modifications. To do so, we exploit the technique of maximal prime subgraph decomposition in determining the minimal subgraph(s) that have to be recompiled, and thereby the minimal subtree(s) of the JT that should be replaced by new subtree(s). We focus on structural modifications: addition and deletion of links and variables.

## 1 INTRODUCTION

Most popular knowledge engineering tools for construction and execution of probabilistic expert systems based on Bayesian networks (BNs), such as for example HUGIN [Hugin] or NETICA [NETICA], work with two representations: 1. A direct representation of the BN as illustrated by the edit window in figure 1, and 2. A computational structure also known as the junction tree or the join tree (JT), illustrated by the run window in figure 1.

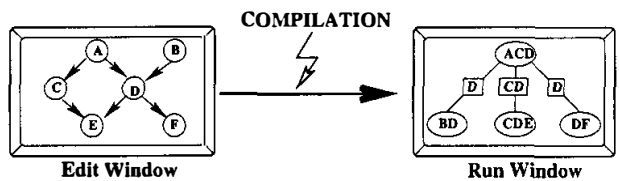

Figure 1: Two representations of a Bayesian network.

A BN is constructed or modified in the first representation and inference is performed over the second representation. Each time the knowledge engineer switches from editing to execution, a *compilation* builds the join tree from scratch. If the network is large, then the CPU time required to perform a new compilation will be considerable. Therefore, it will be desirable that once a join tree representation of the Bayesian network has been generated, incremental changes in the network should produce an update of the previous join tree, and not a new full compilation process. Apart from the *efficiency* reason, there could be more reasons to prefer incremental compilation over total compilation. Draper emphasizes *stability* of the join tree as a desirable property of incremental compilation [Draper, 1995]. It is anticipated that a considerable effort is made to produce an efficient join tree and that incremental changes to an existing (near) optimal join tree will produce more stable results.

Little attention has been directed towards incremental compilation in the literature. Darwiche considers dynamic generation of join trees [Darwiche, 1998], but his aim is to produce efficient elimination orders for specific queries posed to the model rather than modifications over the network. His method is focused on efficient processing by online generation of specific join trees and it does not take modifications of the structure of the underlying network into account. The main goal of Draper [Draper, 1995] is to show how to build join trees avoiding (thinking about) triangulation. To do this, a family graph is initially computed and then transformed into a join tree by using a set of heuristics.

This approach can also be applied in case of dynamic changes in the network structure where the same set of heuristics are applied to dynamically maintain the join tree. A drawback of the method is that it is difficult to identify the relevant set of heuristics and that the resulting join trees are often suboptimal.

In this paper we are primarily interested in structural changes over the network. The method we propose for incremental compilation identifies the parts of the join tree that are affected by changes in the BN, reconstructs only those parts of the join tree and glues the new substructures into the original join tree instead of the outdated parts. This approach ensures a stable and efficient resulting join tree. The method builds on a third representation of the underlying BN the *Maximal Prime subgraph Decomposition (MPD) tree* [Olesen and Madsen, 2002]. A maximal prime subgraph is a subgraph that is d-separated from its surroundings by complete (i.e. fully connected) separators. This enables divide and conquer algorithms for various graph operations such as e.g. triangulations [Olesen and Madsen, 2002] [Flores and Gámez, 2003]. There is a direct correspondence between maximal prime subgraphs (MPS) and subtrees in the join tree, such that a maximal prime subgraph always corresponds to one or more cliques in the join tree. It is this correspondence that is exploited in the present work, by identification of the MPSs that are affected by modifications in the BN. These MPSs determine the parts of the join tree that have to be altered, and an updated join tree can then be constructed incrementally.

In the next section we go through the various issues in incremental compilation. Section 3 summarizes the construction of MPD trees and in section 4 we detail our approach to MPD-based incremental compilation. Finally, we discuss the results and future research in section 5.

## 2 ISSUES IN INCREMENTAL COMPILATION

The starting point in our approach to incremental compilation are changes in an existing BN. Such changes range from simple modifications, e.g. adjustments of numerical parameters in the (conditional) distributions for variables, to complex structural reorganization of variables and their links, possibly altering large parts of the BN. In the following we will go through the possible modifications.

### 2.1 Modification of potentials

The simplest modification of a BN is altering the (conditional) probability distribution for a variable. Such a change is purely quantitative, and it is straightforward as the structure of the join tree will remain unchanged. In this case we simply replace the current table with the modified one.

### 2.2 Modification of the states of a variable

Changes in the state space of a variable can alter the structure of the optimal join tree, because the triangulation typically takes the state space of variables into account. A full treatment of incremental compilation should, of course, consider this class of changes, but the overall efficiency of the resulting computational structure remains roughly the same. We shall therefore disregard the structural implications of such changes from further considerations in the present treatment of the subject. What remains is then the modified structure of the potentials of the altered variable and its children and this is again taken care of by replacing old potentials with new ones for the affected variables.

### 2.3 Removing an arc

Removal of an arc in a BN can be a straightforward change. If, for example, two nodes without parents share a child and is not otherwise connected, removal of an arc between them will not lead to changes in the moral graph, as the link would appear here anyway due to moralization. At the other extreme the removal of an arc could break several cycles in the BN and a considerable simpler join tree could result from a retriangulation of the network. In such cases it is beneficial to be able to identify the minimal part of the join tree that could be affected by the change and concentrate on a retriangulation of only that part.

### 2.4 Adding an arc

As for removal of an arc the addition of a new one is sometimes straightforward. A simple situation is symmetric to the simple case above, where two nodes without parents share a child and is not otherwise connected. Addition of an arc between them will not lead to changes in the moral graph, as the link would appear here anyway due to moralization. At the other extreme the addition of an arc could create several cycles in the BN and in this case large parts of the join tree could be affected. Sometimes a complete retriangulation of the network is required, and again it is beneficial to be able to identify the minimal part of the join tree that could be affected by the change and concentrate on a retriangulation of only that part.

### 2.5 Removing a node

The removal of a node from the BN will include removal of all arcs connected to that node. If all arcs are removed first the removal of the node is simple. If not connected to any other node, the node will constitute an island in the BN and consequently it can simply be deleted.

### 2.6 Adding a node

The addition of a new node is similarly simple, if we connect it to the BN afterwards arc by arc. The node is simply added to the BN and the procedure for adding arcs is called when it is linked with the existing BN.

### 2.7 Incremental compilation

In this paper we are primarily interested in structural changes over the network. We shall therefore concentrate on the addition and deletion of nodes and arcs. The problem we will investigate is illustrated by figure 2. The graph $G$ of the original Bayesian network is shown in the upper left corner and the top arrow illustrates the compilation into the join tree $\mathcal{T}$ in the upper right corner. The left vertical arrow indicates the modification of $G$ into $G'$, and the bottom arrow depicts the approach by existing tools, where $G'$ would be recompiled into $\mathcal{T}'$ in the right lower corner. The approach we suggest is illustrated by the right vertical arrow where $\mathcal{T}'$ is obtained directly from $\mathcal{T}$ based on the changes leading from $G$ to $G'$.

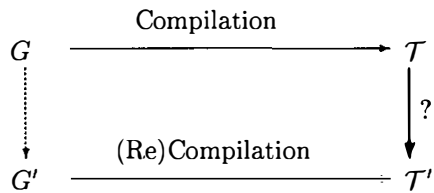

Figure 2: The idea behind incremental compilation.

As mentioned earlier, modifications of the BN can result in everything from trivial to very complex modifications of the join tree. We are therefore looking for ways to limit the parts of the join tree that have to be recompiled. The maximal prime subgraphs of the graph of the BN is the intermediate structure that can resolve this issue. We shall therefore review the method by Olesen and Madsen [Olesen and Madsen, 2002] for identification of a maximal prime subgraph decomposition of a Bayesian network in the following section.

## 3 MAXIMAL PRIME SUBGRAPH DECOMPOSITION

The decomposition of the graph of a Bayesian network into its maximal prime subgraphs is integrated into the well known procedure for construction of join trees for Bayesian networks. We assume the join tree construction procedure known and refer to e.g. Jensen [Jensen, 2001] for details.

The method for identification of the maximal prime subgraphs of the directed acyclic graph (DAG) $G$ of a Bayesian network is based on a junction tree representation $\mathcal{T}$ of $G$. A precondition of the method is that the triangulation $T$ from which $\mathcal{T}$ is constructed is minimal. There exist methods for finding triangulations with a minimal number of fill-in edges such as for example the LEX M algorithm [Rose et al, 1976]. Alternatively, the recursive thinning algorithm by Kjærulff [Kjærulff, 1993] can be applied to remove redundant fill-in edges from an arbitrary triangulation. Usually minimality is not considered important as a minimal triangulation is not necessarily a good triangulation with respect to efficient inference. However, a minimal triangulation $T_{min}$ obtained by recursive thinning of a triangulation $T$ is always at least as good as $T$. The recursive thinning should, of course, only be applied if the triangulation algorithm is not guaranteed to produce minimal triangulations.

The graph, $G$, of a Bayesian network is decomposed into its maximal prime subgraphs as follows. Let $G^M$ be the moral graph of $G$, let $G^{T_{min}}$ be the graph corresponding to a minimal triangulation $T_{min}$ of $G^M$, and let $\mathcal{T}_{min}$ be the corresponding join tree.

The following algorithm describes the construction of a join tree $\mathcal{T}_{min}$:

**Algorithm 1** ConstructJoinTree(DAG G)

1. Moralize $G$ to obtain $G^M$.
2. Triangulate $G^M$ to obtain $G^T$.
3. Thin out redundant fill-in edges to obtain $G^{T_{min}}$.
4. Organize the clique decomposition induced by $G^{T_{min}}$ as a junction tree $\mathcal{T}_{min}$.
5. Return $\mathcal{T}_{min}$.

The maximal prime subgraphs of $G^M$ are formed by aggregating adjacent cliques of $\mathcal{T}_{min}$ connected by a separator which is incomplete in $G^M$. Algorithm ConstructMPDTree performs this process and returns a join tree $\mathcal{T}_{MPD}$, where the nodes represents the maximal prime subgraphs:

**Algorithm 2**
ConstructMPDTree(JoinTree $\mathcal{T}$, Graph $G^M$)

1. Aggregate all adjacent cliques in $\mathcal{T}$ with an incomplete separator in $G^M$ to obtain $\mathcal{T}_{MPD}$.
2. Return $\mathcal{T}_{MPD}$.

Figure 3 illustrates the application of the algorithms on the well-known Asia example by Lauritzen and Spiegelhalter [Lauritzen and Spiegelhalter, 1988]. Part (a) shows the BN for the example and in part (b) the moral graph is obtained by adding arcs between common parents of all nodes (arcs (T, L) and (E, B)), and dropping the directions of the original arcs. In part (c) the triangulated graph results from adding the arc (L, B). The triangulation is minimal and no arcs have to be removed by the recursive thinning step. The resulting join tree is shown in part (d) and a check of the separators in the moral graph yields the maximal prime subgraph decomposition tree shown in part (e). Part (f) gives the maximal prime subgraphs of the Asia network. Remark that this decomposition is unique [Olesen and Madsen, 2002].

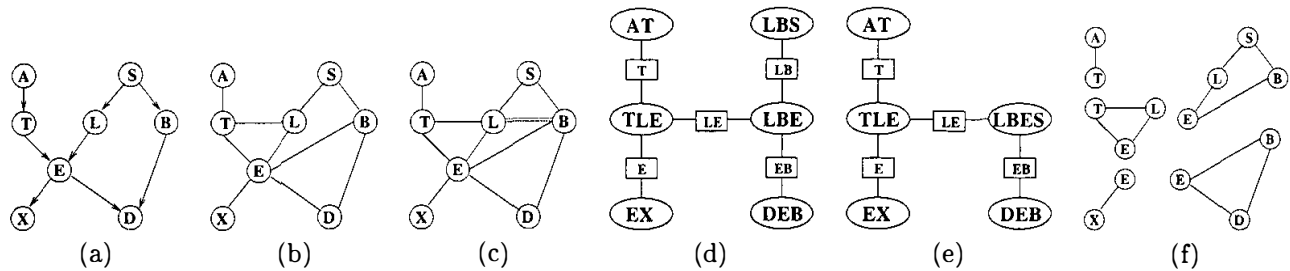

(a)  (b)  (c)  (d)  (e)  (f)

Figure 3: (a) The Asia network. (b) Moral graph for Asia. (c) Triangulated graph for Asia. (d) A join tree for Asia. (e) Maximal Prime Subgraph Tree for Asia. (f) Maximal Prime Subgraph Decomposition for Asia.

As can be seen there is a direct correspondence between MPSs and cliques in the join tree. The structure of the join tree is a refinement of the MPD tree (although constructed in the opposite order), where a node in the MPD tree may be expanded into one or more cliques in the join tree. It is this structural correspondence that is exploited in our method for incremental compilation.

## 4 MPSD-BASED INCREMENTAL COMPILATION

As mentioned in the introduction the maximal prime subgraphs can be triangulated independently. We proceed as follows: Each time a BN is recompiled we identify the set of MPSs affected by the modifications since the last compilation, and only these MPSs will be re-triangulated. Our expectation is that only a few subgraphs of $\mathcal{T}_{MPD}$ will be influenced by the modifications, and consequently only a small part of the graph has to be re-triangulated. Thus, the major part of the join tree remains unmodified and can be reused.

Before explaining in details how to cope with the four basic modifications (remove arc, add arc, remove variable and add variable), we give the general procedure to perform *MPSD-based incremental compilation*. Figure 4 illustrates the process and is used during the explanation provided below.

The process starts with a user making modifications in the edit window of figure 1. The result of this editing process is an updated version of the BN ($G'$) (step (1) in figure 4). A list of the modifications is constructed during this step. This list serves as input to algorithm 3, that is activated when the user decides to obtain a new join tree.

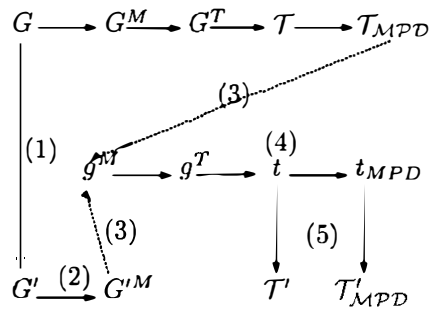

Figure 4: Overview of the MPSD-based incremental compilation process.

Once the user decides to produce a join tree for the new network $G'$, the first step of incremental compilation is to modify the moral graph (Step (2) in figure 4). Notice that the moral graph plays a crucial role in MPSD construction. We will pay special attention to this step in algorithm 4.

As an example, let us suppose that the user has removed link $L \rightarrow E$ and has invoked the IC algorithm. Then, the updated moral graph is the one depicted in part (a) of figure 5. Remark that the moral link T-L has also been removed.

Now, we proceed to step (3) in figure 4. This is the key point in our algorithm, where we identify the minimal set of MPSs which may be affected by the modifications performed over the BN. Shortly we shall detail this for the different modifications, but for the moment, let us assume the existence of an algorithm which marks the MPSs in $\mathcal{T}_{MPD}$ affected by a modification. In the example, this algorithm marks MPSs $(TLE)$ and $(LBES)$ (to be detailed in subsection 4.1). In the example, there is only one connected marked subtree $(TLE) - [LE] - (LBES)$, but in general, when all modifications have been processed by the marking algorithm, there may be more connected parts of $\mathcal{T}_{MPD}$ that have been marked. As an example, consider the scenario in which the marking algorithm marks MPSs: $(AT)$, $(TLE)$ and $(DEB)$; which gives rise to two marked connected subtrees: $(AT) - [T] - (TLE)$ and $(DEB)$. For each of these marked connected subtrees we proceed, in turn, with the following steps.

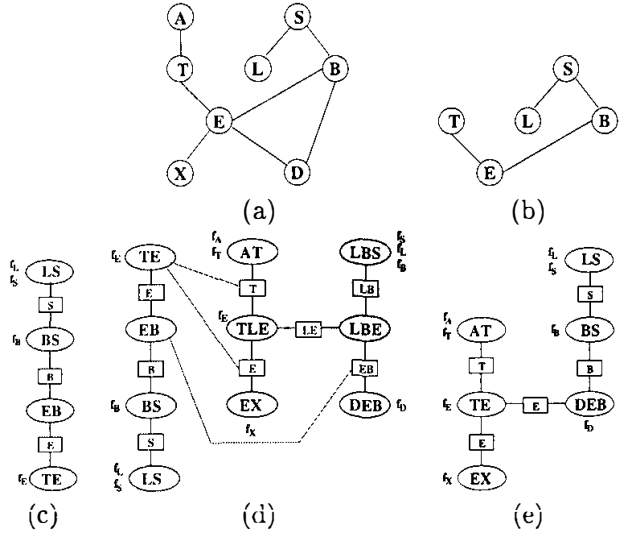

Figure 5: MPSD-based incremental compilation of Asia when removing $L \to E$.

Let $g^M$ be the subgraph of $G'^M$ induced by the set of variables included in a connected marked subtree of $\mathcal{T}_{MPD}$. Figure 5.b shows $g^M$ as the projection of $G^M$ over $\{T, L, E, B, S\}$. In step (4) in figure 4 we obtain a JT and a MPD tree for $g^M$. We obtain a join tree $t$ by applying algorithm 1 (avoiding step 1) and the corresponding MPD tree $t_{MPD}$ by applying algorithm 2. In the example, part (c) of figure 5 shows both structures, because in this case every MPS corresponds uniquely to one clique.

Finally, during step (5) in figure 4 both $\mathcal{T}$ and $\mathcal{T}_{MPD}$ are updated by using the newly obtained structures $t$ and $t_{MPD}$. For each separator $S$ connecting a marked subtree with an unmarked cluster we reconnect $S$ to a cluster of $t$ ($t_{MPD}$) having maximal intersection with $S$. Upon completion, the marked clusters (and separators between them) are deleted. In the example, the separators connecting the marked connected subtree with the rest of $\mathcal{T}_{MPD}$ are $[T], [E]$ and $[EB]$. These separators are connected with the new graphical structure as shown in part (d) of figure 5. Finally, part (e) of the same figure shows the final structure, obtained after removing the outdated part of the tree (the marked MPSs). Algorithm 5 details this procedure.

Algorithm 3 details the incremental compilation process described above. In order to simplify the header of the algorithms presented in this section, we suppose that the main graphical structures are accessible, that is, we will refer to $G^M$, $\mathcal{T}$ and $\mathcal{T}_{MPD}$ without the need of passing them as a parameter to the algorithms. Before we proceed we need some notation. The potential of $X$ is assigned to a specific clique in $\mathcal{T}$, which contains the family of $X$. In the following, we will use $C_X$ to identify this clique and, likewise, $M_X$ will identify the MPS in $\mathcal{T}_{MPD}$ which has the family of $X$ associated. In figures this will be indicated with a $f_X$ next to the corresponding clique/MPS.

The first loop of algorithm 3 iterates over all modifications. For each modification we adjust the moral graph by algorithm 4. Algorithm 4 maintains a list of all links that are affected by the actual modification. This is relevant for addition and deletion of arcs, and algorithm 4 returns a list, $L$, of the added (deleted) link and appends induced added (deleted) moral links.

**Algorithm 3**
IncrementalCompilation (Modification list $ModList$)

1. For each modification $mod$ in $ModList$ do
   (a) $L \leftarrow$ ModifyMoralGraph($mod$)
   (b) Case $mod$ of
       i. Add node $X$: AddNode($X$)
       ii. Delete node $X$: RemoveNode($X,M_X,nil$)
       iii. Delete link $X \to Y$: RemoveLink($L,M_Y,nil$)
       iv. Add link $X \to Y$: AddLink($L$)
2. For each connected marked subtree, $\mathcal{T}_{MPD}$, of $\mathcal{T}_{MPD}$ do
   (a) Mark all cliques in the subtree $T$ of $\mathcal{T}$ corresponding to $\mathcal{T}_{MPD}$
   (b) Let $C$ be any cluster of $T$ and let $M$ be any cluster of $\mathcal{T}_{MPD}$
   (c) $V \leftarrow \{$all variables included in $\mathcal{T}_{MPD}\}$
   (d) $g^M \leftarrow G^M(V)$
   (e) $t \leftarrow$ ConstructJoinTree($g^M$)
   (f) $t_{MPD} \leftarrow$ AggregateCliques($t$)
   (g) $\mathcal{T} \leftarrow$ connect($t, C, nil$)
   (h) $\mathcal{T}_{MPD} \leftarrow$ connect($t_{MPD}, M, nil$)
   (i) Delete $T$ and $\mathcal{T}_{MPD}$

**Algorithm 4** ModifyMoralGraph(Modification $mod$)
1. $L \leftarrow \emptyset$　　　// a link list
2. Case $mod$ of
   i. Add node $X$: add a new (isolated) node $X$ to $G^M$
   ii. Delete node $X$: remove $X$ from $G^M$
   iii. Add link $X \to Y$: add $X \to Y$ to $L$ together with all new links needed to make $Y \cup parents(Y)$ a complete sub-graph
   iv. Delete link $X \to Y$:
       (a) if $(children(X) \cap children(Y) = \emptyset)$ then
           - delete $(X,Y)$ from $G^M$
           - add $(X,Y)$ to $L$
       (b) for all $Z_i \in parents(Y) \setminus \{X\}$ do
           - if $((children(Z_i) \cap children(X) = \{Y\})$ and $(Z_i \to X$ or $X \to Z_i)$ not in $G$) then
             - delete $(X, Z_i)$ from $G^M$
             - add $(X, Z_i)$ to $L$
3. Return $L$

The list, $L$, returned by algorithm 4 is passed on as argument to the relevant procedure, that marks affected MPDs in $\mathcal{T}_{MPD}$. This result of step 1 in algorithm 3 is an MPD-tree with (possible several) connected marked subtrees. In the second step of algorithm 3 we iterate over these subtrees and adjusts $\mathcal{T}$ and $\mathcal{T}_{MPD}$ by algorithm 5. The pattern for this algorithm may not be immediately transparent. The recursive control structure acts on the two last parameters where the former (the second parameter) is the cluster to which the algorithm is applied, and the latter (the third parameter) is the caller. The structure traverses a marked subtree, avoiding loops by a check that the caller is not re-visited. This pattern is also used in algorithms 6 and 7.

**Algorithm 5**
Connect(Cluster Tree $t$, Cluster $C_i$, Cluster $C_j$)

For each separator $S$ between $C_i$ and $C_k \neq C_j$ do

If $C_k$ is unmarked then
(a) locate cluster $C \in t$ such that $C \cap C_k$ is maximal
(b) Connect $C$ with $C_k$ by $S$
(c) If $S = C$ then amalgamate $C$ and $C_k$

else connect($t$, $C_k$, $C_i$)

We shall now go through the details of the marking of MPDs, that is, the procedures called from the first loop of algorithm 3. It is silently assumed that whenever a MPD is marked, the corresponding cliques in the join tree will also be marked.

### 4.1 Removing a link

Let us suppose that link $X \to Y$ has been deleted from $G$. Then $M_Y$ has been affected and we have to investigate if more MPSs have to be re-triangulated due to a side effect of the deletion of $X \to Y$. Therefore, we should include neighbors of $M_Y$ in the set of MPSs to re-triangulate, only if the separator between them is no longer complete. To do this, we look if the disappearance of the link $X - Y$ or of any other induced link $Z - X$ causes some separator to become incomplete in $G'^M$. Of course, if a new MPS is marked because of this search, then we have to verify the same condition among their neighbors and so on.

The following algorithm marks the MPSs affected by the removal of $X \to Y$. Parameter $L$ is the list of (induced) moral links (returned by ModifyMoralGraph). Notice, that when the algorithm is called the first time (from Incremental Compilation) then $M_Z = nil$.

As a case of study, let us to retake the example used during the overview of our IC method, that is, the removing of link $L \to E$ from the Asia network. In this case, the parameters received by algorithm 6 are $L = \{(L,E),(T,L)\}$, $M_Y = (TLE)$ and $M_Z = nil$.

Therefore, MPS $(TLE)$ is marked in step 1. From the three separators connected to $(TLE)$, only $[LE]$ will be considered, because the other two contain only one variable. As $[LE]$ contains one of the removed links, the MPS $(LBES)$ connected to $(TLE)$ by this separator, is also marked by a recursive call of removeLink. Therefore, in this case, the subtree $(TLE) - [LE] - (LBES)$ is marked by the algorithm.

**Algorithm 6**
RemoveLink(LinkList $L$,MPS $M_Y$,MPS $M_Z$)

1. Mark $M_Y$
2. For all neighbors $M_K \neq M_Z$ of $M_Y$ do
   (a) $S \leftarrow$ separator between $M_Y$ and $M_K$
   (b) if $L \cap links(S) \neq \emptyset$ then RemoveLink($M_K, M_Y, L$)

### 4.2 Removing a node

Algorithm 7 marks all MPSs containing $X$. The algorithm also deletes $X$ from all MPSs and separators containing it in order to obtain the correct set $V$ in step (c) of the second loop of algorithm 3.

**Algorithm 7**
RemoveNode(Node $X$, MPS $M_X$, MPS $M_Y$)

1. Delete $X$ from $M_X$ and mark $M_X$
2. For all neighbors $M_Z \neq M_Y$ of $M_X$ do
   (a) $S$ be the separator between $M_X$ and $M_Z$
   (b) if $X \in S$ then
       i. Delete $X$ from $S$
       ii. RemoveNode($X,M_Z, M_X$)

As an example, let us consider the removal of variable $D$ from the Asia network. This operation results in the following list of modifications: (remove $E \to D$, remove $B \to D$, remove $D$), which is passed to alg. 3 from the edit mode. Subsequently, the moral link $(E \to D)$ is removed during the first loop of algorithm 3. In the second loop of algorithm 3 the MPSs $(DEB)$ and $(LBES)$ are marked. In this case, the effect of algorithm 7 is just to remove $D$ from $(DEB)$. Therefore, we have to re-triangulate $G'^M(\{E,B,L,S\})$, see figure 6(a). Part (b) of the same figure shows the obtained trees from this graph. In part (c) the connecting process of the old and new structure is shown, where marked clusters are highlighted by filling them. Finally fig. 6(d) shows the obtained result after absorbing non maximal cluster $(LE)$ into cluster $(TLE)$.

### 4.3 Adding a node

This a very simple operation. As $X$ is a new variable, it will be an isolated node in the network, so, the modification consists of the addition of a new MPS/clique containing only variable $X$. In order to maintain single structures (trees) rather than sets (forests), we connect new clusters to the respective trees by picking up any existing cluster and connect the new cluster by an empty separator. The process is detailed in algorithm 8.

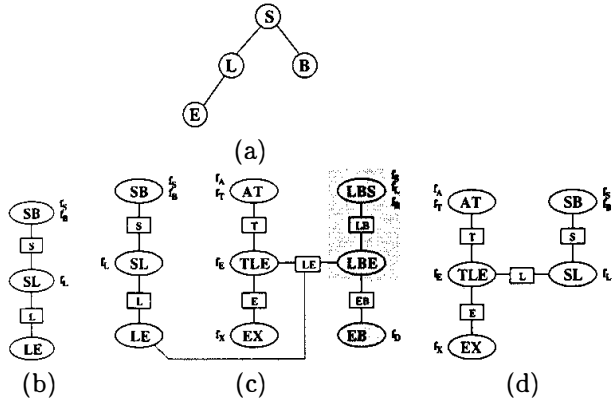

(a)　(b)　(c)　(d)

Figure 6: MPSD-based incremental compilation of Asia when removing variable $D$.

**Algorithm 8** AddNode(Node $X$)

1. Create a new marked clique $C_X$ and a new marked MPS $M_X$ containing only $X$
2. Connect $C_X$ to $\mathcal{T}$ by an empty separator
3. Connect $M_X$ to $\mathcal{T}_{MPD}$ by an empty separator

As an example, the result of adding a new variable $Z$ to the Asia network is shown in figure 7.

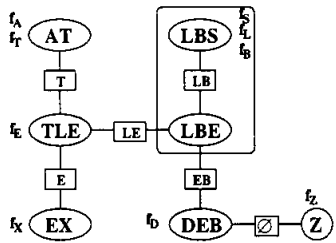

Figure 7: MPSD-based incremental compilation of Asia when adding a new variable $Z$.

### 4.4 Adding a link

Finally, we will consider the addition of a new arc $X \to Y$. This change will (at least) modify $M_Y$. If $X$ is already included in $M_Y$, then only this MPS has to be re-triangulated. Otherwise, we have to look for a MPS, $M_X$, in which $X$ is included ($M_X$ is not necessarily the MPS to which $X$ originally was assigned). $M_X$ and $M_Y$ are marked and so are all the MPSs on the path between them.

This is the general idea of the method for adding a link, however, there is a tricky point that should be discussed. Due to the presence of empty separators, it is possible to modify the tree structure after having located $M_X$, in order to achieve a better (more efficient) structure for our goal. For example if $A \to Z$ is the link to be added to the structure in figure 7, then we will mark all the MPSs in the tree except $(EX)$. However, as $M_Z$ is connected to the tree by an empty separator, we can connect it to MPS $(AT)$ instead. By using this new tree, only MPSs $\{(AT),(Z)\}$ have to be re-triangulated, which leads to a (far) more efficient process.

Algorithm 9 marks the MPSs affected by the addition of $X \to Y$.

**Algorithm 9** AddLink(LinkList $L$)

For each Link $X \to Y$ in $L$ do

1. Let $M_X$ be the nearest neighbor to $M_Y$ containing $X$.
2. If there is an empty separator $S$ on the path between $M_X$ and $M_Y$ then

    (a) Disconnect $\mathcal{T}_{MPD}$ and delete $S$
    (b) Connect $M_X$ to $M_Y$ by an (artificial) separator containing $X$

3. Mark $M_X$, $M_Y$ and all $M_Z$ on the path between them

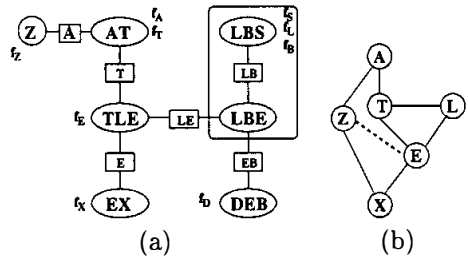

(a)　(b)

Figure 8: Incremental compilation of the structure in figure 7 when adding links $\{A \to Z, Z \to X\}$.

As an example, let us to consider the addition of two new links: $A \to Z$ and $Z \to X$ to the structure depicted in figure 7:

Adding $A \to Z$: as there is an empty separator in the path between $(Z)$ and $(AT)$, the tree is modified to the one depicted in figure 8(a). Algorithm 9 marks MPSs $(Z)$ and $(AT)$. Also, the separator is set to $A$.

Adding $Z \to X$: Now, there is no empty separator along the path between $(Z)$ and $(EX)$, so no modification is performed over the tree. The algorithm marks $(Z),(AT),(TLE)$ and $(EX)$ as the MPSs to be re-triangulated.

Remark that the moral link $Z - E$ has been added to $G'^M$ (by algorithm 4).

We get $\{(Z),(AT),(TLE),(EX)\}$ as the set of MPSs to be re-triangulated. The subgraph of $G'^M$ induced by the set of variables in these MPSs ($\{Z,A,T,L,E,X\}$) yields the graph shown in part (b) of figure 8, and re-triangulating this we get the tree depicted in part (a) of figure 9. Finally, the connecting process is illustrated in fig. 9(b) and the final result

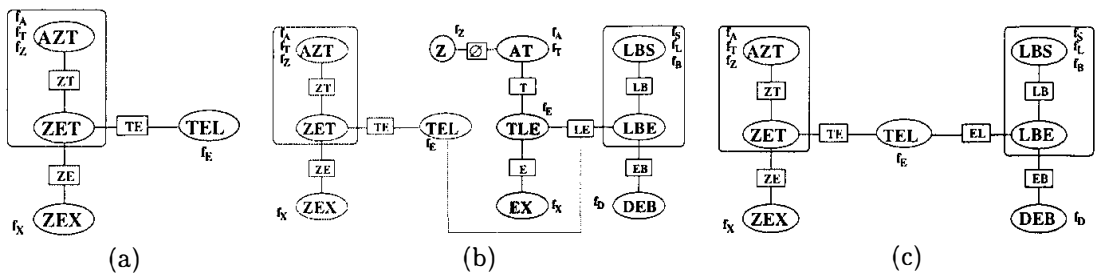

Figure 9: Continuation of example in figure 8.

(after removing marked clusters) is shown in part (c) of the same figure.

## 5 CONCLUDING REMARKS

We have presented a method for incremental compilation of Bayesian networks. The method depends on a maximal prime subgraph representation of the graph of the Bayesian network, which is easily obtainable from the join tree representation. The key point is that the maximal prime subgraphs are the minimal subgraphs that can be triangulated independently, and the method thereby ensures that no superflous computations are carried out. Moreover, the method supports stability of the join tree as existing parts are recycled and only the parts that have been affected by changes in the BN are modified.

The method saves time in situations with frequently changing BNs. This is typically the case during construction and tuning of models, and especially during learning of BN models minor modifications are often systematically applied. In such processes huge model spaces are searched and modifications will most often consist of addition or removal of a single arc.

The main algorithm IncrementalCompilation is constructed such that it can be activated at any time. This enables the procedure to operate both in simple mode, where it is called for each simple modification to the BN, and in batch mode, where several modifications are processed simultaneously. Thus, a situation similar to inference, where inference can be performed either for each piece of evidence or for a group of findings, can be obtained. This feature is seen in most tools for construction and execution of BN models.

We are currently implementing the methods in order to investigate the practical impact, but we are confident that it will prove to save time in several practical situations.

In the present paper we have restricted ourselves to discrete Bayesian networks and a topic for further research is to investigate the approach for other kinds of graphical models, such as CG-models and influence diagram.

### Acknowledgements
The work has been partially supported by the Spanish Ministerio de Ciencia y Tecnología (MCyT), under project TIC2001-2973-C05-05.